\documentclass[letterpaper, 10 pt, conference]{ieeeconf}  

\usepackage{caption}
\usepackage{subcaption}
\usepackage{algpseudocode}
\usepackage{algorithm}
\usepackage{graphicx}
\usepackage{amsmath}
\usepackage{multirow}
\usepackage{tabularx}
\usepackage{svg}
\usepackage{bbm}
\usepackage{hyperref} 
\usepackage{cleveref} 
\usepackage{cite} 

\IEEEoverridecommandlockouts 
\title{\LARGE \bf WiFi-CSI Sensing and Bearing Estimation in Multi-Robot Systems: \\An Open-Source Simulation Framework}

\author{Brendan Dijkstra$^{1}$, Ninad Jadhav$^{2}$, Alex Sloot$^{1}$, Matteo Marcantoni$^{1}$, Bayu Jayawardhana$^{1}$, Stephanie Gil$^{2}$, \\and Bahar Haghighat$^{1}$
\thanks{$^{1}$B. Dijkstra, A. Sloot, M. Marcantoni, B. Jayawardhana, and B. Haghighat are with the University of Groningen, 9747 AG Groningen, The Netherlands
        {\tt\small b.j.dijkstra.2@student.rug.nl}, {\tt\small a.h.sloot@student.rug.nl}, {\tt\small m.marcantoni@rug.nl}, {\tt\small b.jayawardhana@rug.nl},  {\tt\small bahar.haghighat@rug.nl}}%
\thanks{$^{2}$N. Jadhav and Stephanie Gil are with the Harvard University, Cambridge, MA 02138, USA
        {\tt\small njadhav@g.harvard.edu},  {\tt\small sgil@seas.harvard.edu}}%
}

\begin{document}

\maketitle
\thispagestyle{empty}
\pagestyle{empty}

\def\arraystretch{1.15}

\begin{abstract}

Development and testing of multi-robot systems employing wireless signal-based sensing requires access to suitable hardware, such as channel monitoring WiFi transceivers, which can pose significant limitations. The WiFi Sensor for Robotics (WSR) toolbox, introduced by Jadhav et al. in 2022, provides a novel solution by using WiFi Channel State Information (CSI) to compute relative bearing between robots. The toolbox leverages the amplitude and phase of WiFi signals and creates virtual antenna arrays by exploiting the motion of mobile robots, eliminating the need for physical antenna arrays. However, the WSR toolbox's reliance on an obsoleting WiFi transceiver hardware has limited its operability and accessibility, hindering broader application and development of relevant tools.
We present an open-source simulation framework that replicates the WSR toolbox's capabilities using Gazebo and Matlab. By simulating WiFi-CSI data collection, our framework emulates the behavior of mobile robots equipped with the WSR toolbox, enabling precise bearing estimation without physical hardware. We validate the framework through experiments with both simulated and real Turtlebot3 robots, showing a close match between the obtained CSI data and the resulting bearing estimates.
This work provides a virtual environment for developing and testing WiFi-CSI-based multi-robot localization without relying on physical hardware. All code and experimental setup information are publicly available at \href{https://github.com/BrendanxP/CSI-Simulation-Framework}{https://github.com/BrendanxP/CSI-Simulation-Framework}.

\end{abstract}

\setlength{\belowcaptionskip}{-3pt}

\section{Introduction}
\label{sec:intro}
Development and testing of multi-robot systems using WiFi-based sensing technologies face significant challenges due to limited hardware availability. Real-world testing requires specialized WiFi transceivers which require customized drivers and quickly become outdated. This limits the ability of researchers to develop and experiment with new algorithms that leverage WiFi Channel State Information (CSI) for tasks such as multi-robot localization and coordination. 
This work aims to lower the barrier to WiFi-based multi-robot localization research based on WiFi CSI data by providing an open-source simulation framework, eliminating the need for configuring and deploying hardware setups.

\begin{figure}[t!]
    \centering
    \includegraphics[width = 1.0\columnwidth]{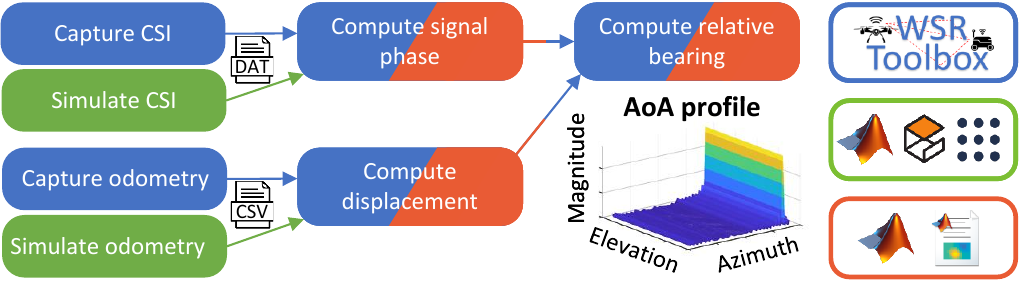}

    \caption{Overview of the operation of our simulation framework (green and orange blocks) in parallel with the WSR toolbox (blue blocks) \cite{Jadhav2022ToolboxRobotics}. The blocks in green involve implementations in Matlab, Gazebo, and ROS. Those in orange involve Matlab. The WSR blocks in blue involve implementations in C++ and Python \cite{Jadhav2022ToolboxRobotics}. Three steps are included in our framework: (i) simulating raw CSI and robot odometry data, (ii) computing phase using CSI data, and (iii) running Bartlett's estimator to obtain robots' relative bearing.
    }
    \label{img:tools_overview}
\end{figure}

Warehouse automation trends have been largely pushing the demand for affordable, accurate, and easily deployable multi-robot systems \cite{Arents2022SmartManufacturing, Queralta2020CollaborativeVision, Tantawi2019AdvancesCollaboration, Chakraborti2020FromForum}. Indoor localization is a central requirement for successful deployment of these systems \cite{Cadena2016PastAge, Burgard2000CollaborativeExploration, Stump2011Multi-robotProblem, Chen2015AdaptiveVision, Fenwick2002CooperativeLocalization, Jadhav2022ToolboxRobotics}. 
Indoor deployments often rely on vision systems, wireless beacons or Ultra-Wide Band (UWB) signals, which can be costly and time-consuming to setup \cite{Prorok2013AccurateCollaboration, Martinelli2020CooperativeCases}. Leveraging an existing Radio Frequency (RF) infrastructure, such as WiFi, offers a viable alternative for indoor localization due to its cost-effectiveness and widespread availability.

While UWB has gained considerable attention for its accuracy in localization, the associated regulatory restrictions and high setup costs complicate widespread deployment \cite{Jadhav2022ARobotics, Mahfouz2009RecentPositioning, 2019COMMISSIONRepealing}. In contrast, WiFi-based localization offers a widely available, low-cost alternative for multi-robot systems. WiFi signals can operate in both line-of-sight and non-line-of-sight (NLOS) conditions, making them ideal for indoor environments where GPS signals are inaccessible. Additionally, WiFi CSI provides amplitude and phase data, enabling precise relative range \cite{Vasisht2016Decimeter-LevelPointb} and bearing \cite{Jadhav2022ARobotics} estimation. 

The WiFi CSI data, contains information about WiFi signals carrier channel and the signal propagation paths between transceiving nodes.
CSI data has been used in a wide variety of sensing tasks, such as imaging and tracking through obstacles \cite{Adib2015CapturingWall,Depatla2017RoboticImaging,Adib2013SeeWi-Fi}, shape and activity detection \cite{Jin2018WiSh:RFIDs,CominelliExposingLimitations,Wang2019AInformation}, and material sensing \cite{Zhang2019OnSensing,Ha2020FoodRFIDs,Dhekne2018TrackIO:Inside-Out}. Despite the wide applicability, the existing methods often overlook the challenges and opportunities posed by mobility of transceiver nodes, limiting their integration with robotic platforms. Moreover, these methods make use of a wide range of different specialized hardware and software for capturing and processing CSI data and lack a standard framework for simulating realistic CSI.

Conventional approaches to WiFi-based localization leverage Received Signal Strength Information (RSSI), which quantifies the magnitude of the incoming signal \cite{Zhu2013RSSI-basedLocalization, Coppola2018On-boardTeams, Amoolya2019Wi-FiEnvironment}. However, this method has limited sensitivity (displacements larger than 1m are typically required) and is prone to noise \cite{Jadhav2022ToolboxRobotics,Zafari2019ATechnologies, Yan2013Co-OptimizationEnvironments}.
Modern WiFi Network Interface Cards (NICs) using 802.11n/ac/ax standards, also known as WiFi 4 and up, capture CSI to optimize signal transmission in the NICs internal operations \cite{IEEEStandardsAssociation2023TheStandards,Wi-FiAlliance}. CSI data contains characteristic information about the wireless channel and the propagation paths, enabling an analysis of environment properties \cite{Jadhav2022ARobotics}. This information essentially provides an estimation of the real Channel Frequency Response (CFR) $H(t)$ in \Cref{eq:cfr}, where $X(t)$ and $Y(t)$ are the transmitted and received signals, respectively \cite{Liu2018WiCount:Signals,Wang2019AInformation}.
\begin{equation}
    Y(t)=H(t)X(t)+N
    \label{eq:cfr}
\end{equation}

Custom drivers (e.g., for the Intel WL5300 on the WiFi Sensor for Robotics (WSR) toolbox \cite{Jadhav2022ToolboxRobotics,Halperin2011ToolRelease}) or kernel-level code changes tailored to specific NICs (e.g. the Nexmon Patching Framework on a Raspberry Pi 4B \cite{Schulz2016NexmonHackt}) are needed to obtain the CSI data captured by NICs for any custom developments. 
This paper contributes an open-source realistic physics-based simulation framework as shown in \Cref{img:tools_overview} that replicates the operation of mobile robots equipped with the hardware and software systems of the WSR toolbox, facilitating planning experiments and studying scenarios where the WSR toolbox is deployed.

\section{The WSR Toolbox}

The WSR Toolbox published in 2022 is a novel, proven, and open-source solution to obtaining WiFi-CSI-based bearing estimations on moving robots \cite{Jadhav2022ARobotics}. The toolbox enables real-time WiFi-CSI-based localization, rendezvous, or mapping purposes. It performs the hardware and software operations onboard, in real-time, and scalable for multi-robot systems \cite{Jadhav2022ARobotics,Jadhav2022ToolboxRobotics}. By leveraging the robot's motion, WSR toolbox captures WiFi-CSI data using a virtual antenna array.

\begin{figure}[t]
    \centering
    \includegraphics[width=\linewidth]{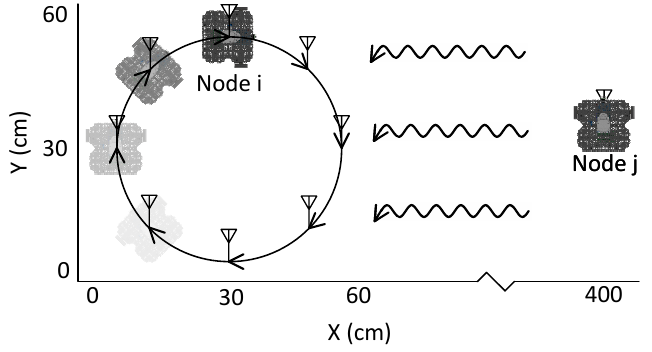}
    \caption{Creating a virtual antenna array by moving a single antenna element on a mobile robot in a SAR-based approach. This example shows a circular array with a 30 cm radius and a transmitting node 4m apart to establish the far-field assumption considered in the WSR toolbox developments.}
    \label{fig:antenna-array}
\end{figure}

The hardware setup is based on an Intel WL5300 NIC that is operated using customized drivers to extract the CSI data. This NIC is installed on a single-board computer (SBC) that is mounted on ground or aerial robots. 
The WSR Toolbox offers a unique set of tools for creating a virtual antenna array using a single antenna element in a Synthetic Aperture Radar (SAR)-based approach. In this approach, a single antenna is mounted on a mobile robots and captures multiple different signals while traversing a trajectory, as shown in \Cref{fig:antenna-array}. This differs from a physical antenna array where all of the antenna elements can capture a single transmitted signal, in that multiple transmissions are required. However, both approaches can be leveraged to effectively capture the signal phase at multiple different locations. In the virtual antenna array setup, the robotic hardware needs to capture accurate odometry to compute the exact geometry of the resulting virtual antenna array resulting from its motion \cite{Jadhav2022ARobotics}. 

The transmitting nodes require an efficient method for transmitting and receiving packets required for capturing the CSI information, in order not to saturate the channel bandwidth. The WSR toolbox employs a round-robin protocol similar to the Time-Division Multiple Access (TDMA) algorithm, where each node $j$ only transmits a packet after it receives a packet from node $i$ that was specifically intended for node $j$. 
To cancel noise, the toolbox utilizes a pair of forward and backward packets that are broadcast almost simultaneously \cite{Jadhav2022ARobotics}. Each packet is therefore given a frame number, to couple the data from the different nodes during processing, lowering the reliance on accurately assigned timestamps. 
The custom driver in the WSR toolbox exports 30 of the 48 subcarriers, which during processing get interpolated to the central subcarrier of the channel \cite{Sen2012YouInformation,Jadhav2022ToolboxRobotics}.

Our simulation framework is shown along with the WSR toolbox in \Cref{img:tools_overview}. After the capture of CSI and odometry data, these data streams get parsed and processed to obtain the WiFi phase data and robot’s displacement. This data is combined to generate the Angle-of-Arrival (AOA) profile, which will be discussed in detail in \Cref{subsec:bearing}.
The AoA profile is a 2D matrix representing the relative paths that a wireless signal takes between a transmitting and a receiving node, across all possible incoming directions of the signal. These directions consist of the azimuth angle in 2D and also contain the elevation in 3D experiments. The strongest signal direction, known as \(AoA_{max}\), is then considered as the final estimated bearing angle between the nodes and can be given a likelihood score based on the AoA profile magnitude at \(AoA_{max}\). The WSR toolbox thus controls the coordinated packet transmission and captures all processing steps leading to the AoA profile with the final bearing estimate \cite{Jadhav2022ARobotics}.

Despite its versatility, operating the WSR toolbox has been limited due to the obsoleting required WiFi NIC and the complexity of the corresponding software. The WL5300 NIC limits Ubuntu usage up to version 18.04 with Linux Kernel 4.15, which has been End-Of-Life (EOL) since mid 2018. Moreover, this version of Ubuntu is no longer actively maintained and is incompatible with recent software tools.


\section{Proposed simulation framework} \label{sec:sim_framework}
\label{sec:sim_tools}

Our proposed simulation framework operates in parallel to the WSR Toolbox in three distinct steps: (i) obtaining raw CSI and robot odometry data, (ii) computing signal phase and robot displacement data, and (iii) running Bartlett's estimator to compute relative bearing.
We develop a new set of tools using MATLAB, ROS1, and Gazebo Classic, as shown in \Cref{img:tools_overview}. ROS allows MATLAB to interface directly with the simulated nodes in Gazebo and simulate the CSI in real-time. We opted for the use of Gazebo given its tight integration with ROS, however, any robotic simulator that operates over ROS can be incorporated into this framework. 
This section will cover how the CSI is simulated and provide comprehensive details on a major noise factor which is the Carrier Frequency Offset (CFO). The bearing estimation using Bartlett's estimator is explained as well as the software pipeline that provides the backbone to this framework.

The CSI received at node \( i \) and broadcast by node \( j \) is denoted as \( h_{ij}(t) \) and shown in \Cref{eq:ideal-channel-signal} \cite{Tse2005FundamentalsCommunication}. We model the WiFi carrier wave as a monochromatic electromagnetic wave, given its constant radio frequency. The carrier wave equation is based on the signal wavelength \( \lambda \) and the Euclidean distance \( d_{ij}(t) \) between the transceiving nodes.
It is important to note that \( d_{ij} = d_{ji} \) and is therefore always written as \( d_{ij} \). The span of time between \( t=t_0 \) to \( t=t_{end} \) encompasses the entire duration of the robot trajectory that is used to form the geometry of the virtual antenna array.

\begin{equation}
    h_{ij}(t) = \frac{1}{d_{ij}(t)}e^{-2\pi i\frac{d_{ij}(t)}{\lambda}}
    \label{eq:ideal-channel-signal}
\end{equation}

\subsection{Carrier Frequency Offset (CFO)} \label{subsec:cfo}

In real-world scenarios, the received signal phase is affected by Carrier Frequency Offset (CFO), which arises due to slight deviations in the configured carrier frequency of the transceiving antennas \cite{Jadhav2022ToolboxRobotics,Sen2012YouInformation}.
In our framework, we model the CSI from node $i$ to $j$ impacted by CFO as $\hat{h}_{ij}(t)$, as shown in \Cref{eq:cfo-channel-signal} \cite{Liu2018WiCount:Signals,Wang2019AInformation,Sen2012YouInformation,Jadhav2022ARobotics,Jadhav2022ToolboxRobotics}. 

\begin{equation}
\begin{split}
    \hat{h}_{ij}(t) &= \frac{1}{d_{ij}(t)} e^{-2\pi i ( \frac{d_{ij}(t)}{\lambda} \hspace{0.04cm} + \hspace{0.06cm} \textit{CFO}_{ij}(t) \hspace{0.04cm} + \hspace{0.04cm} N_i )}\\
    \hat{h}_{j i}(t) &= \frac{1}{d_{ij}(t)} e^{-2\pi i ( \frac{d_{ij}(t)}{\lambda} \hspace{0.04cm} - \hspace{0.06cm} \textit{CFO}_{ij}(t) \hspace{0.04cm} + \hspace{0.04cm} N_j )}    
    \label{eq:cfo-channel-signal}
\end{split}
\end{equation}

Our model of the CSI contains a time-varying phase offset $\textit{CFO}_{ij}(t)$ and a noise $N$. The CFO-based phase offset is modeled as equal components but with opposing signs for $\hat{h}_{ij}(t)$ and $\hat{h}_{ji}(t)$ based on the principle of channel reciprocity. This offset can be removed by multiplying the signals received at the two nodes.

\begin{equation}
    h_{ij}^2(t)
    = h_{ji}^2(t)
    = \hat{h}_{ij}(t) \cdot \hat{h}_{ji}(t) 
    \label{eq:csi-multiplied}
\end{equation}

\begin{figure}[t]
\centering
\begin{subfigure}{0.64\columnwidth}
  \centering
  \includegraphics[width=1\linewidth]{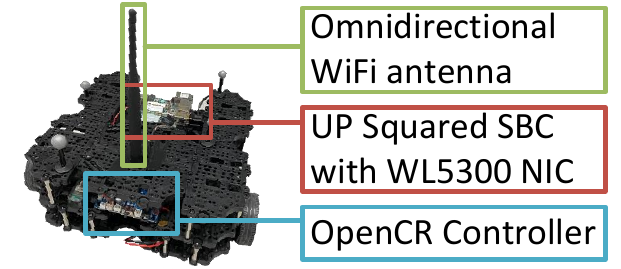}
  \caption{Real robot with onboard WiFi module.}
  \label{fig:robot-real}
\end{subfigure}%
\hfill
\begin{subfigure}{0.32\columnwidth}
  \centering
  \includegraphics[width=.9\linewidth]{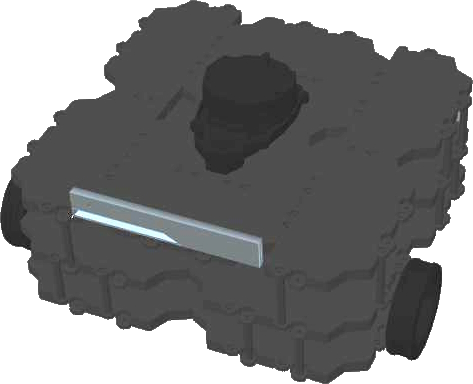}
  \caption{Simulated robot.}
  \label{fig:robot-sim}
\end{subfigure}
\caption{We use Turtlebot3 Waffle robots in our simulation and real setups to collect data. (a) Real robot with onboard UP Squared SBC, Intel WL5300 NIC, and WiFi antenna. (b) Simulated robot with the WiFi antenna at the model's center.}
\label{fig:robot}
\end{figure}

\subsection{Simulating the CSI data} \label{subsec:csi_sim}
We replicate the most prominent factors in the CSI phase by referring to the data that we capture in real experiments, given that phase is used in our bearing estimation later on. These include the CFO, the Sampling Time Offset (STO), and the White Gaussian Noise (WGN).

The sampling time does not consist of precise discrete intervals but contains small time deviations, termed Sampling Time Offset (STO). The STO across all our real experiments measures on average 300 $\mu$s, with an expected sampling period of 10 ms. 
The STO is applied in a stochastic manner to the CSI in \Cref{eq:cfo-channel-signal} in the simulation framework, creating distinct CSI data for each node.

Next, we observe noise in the CSI phase after the CFO cancellation. This can be caused by a plethora of factors, including but not limited to, antenna inconsistencies, signal interference, or changing signal multi-path. We assume this noise to be stochastic WGN and capture it using the Signal-to-Noise Ratio (SNR) by comparing the real CSI phase to the theoretical signal phase. The SNR differed between experiments but was found to be around 3 dB, which is applied in the simulation framework to both nodes.

Lastly, we add two components to model CFO, a constant and a time-dependent component. The constant part consists of $\Delta_f$ observed in the real data, and the variable part is modeled as a sinusoidal that best fits the observed real data.

\subsection{Estimating relative bearing}\label{subsec:bearing}
\begin{figure}
    \centering
    \includegraphics[width=1\linewidth]{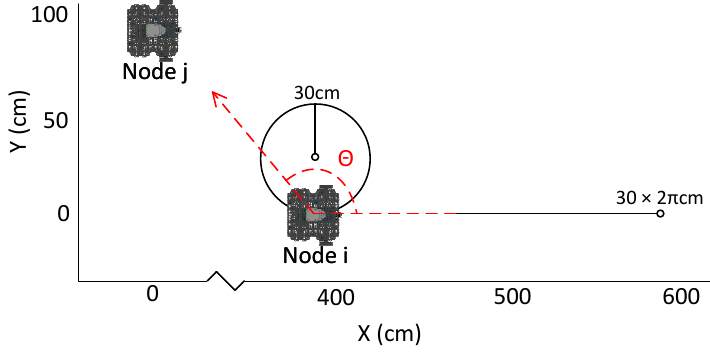}
    \caption{In our experiment setup we use one static robot and one moving robot traversing multiple linear and circular trajectories. The azimuth angle $\theta$ between the two robots is shown, measured with respect to robot $i$'s reference frame.}
    \label{fig:experiment-setup}
\end{figure}

An antenna array allows for capturing an incoming radio signal at multiple different locations. Moving a single antenna to multiple locations can create a virtual antenna array. This, however, requires multiple WiFi transmissions in a Synthetic Aperture Radar (SAR)-based approach \cite{Fitch1988SyntheticRadar, Wiley1985AEvolution}. The transmission problem can be simplified if we consider distances $d_{ij}$ that are much larger than the size of the antenna array, which constitutes the receiving robot's trajectory. This far-field assumption enables us to approximate the incoming signal waves as planer waves. Each element of the antenna array receives the signal from approximately the same direction, as shown in the experiment setup depicted in \Cref{fig:experiment-setup}.

The Angle of Arrival (AoA) profile $F_{i j}(t)$ contains information on the direction of the incoming radio signals and captures the wireless signal's relative paths from robot $i$ to robot $j$. In 2D space, the AoA profile consists of the azimuth angle $\theta$ on the interval $[-180,180]$ degrees. In 3D space, the AoA profile also includes the elevation angle $\phi$, which ranges in $[-90,90]$ degrees. The AoA profile can be calculated with varying precision. We use 1-degree intervals similar to \cite{ Jadhav2022ARobotics,Jadhav2022ToolboxRobotics}. Considering the far-field assumption and the relative displacement required to create a virtual antenna array, a higher resolution often will not yield much more accuracy. 

As input to the Bartlett's estimator for obtaining the relative bearing, we require information about the shape of the virtual antenna array \cite{Stoica2005SpectralSignals}. The steering vector contains the carrier signal's phase changes over the elements of the virtual antenna array. We define the steering vector as $\mathbf{a}(\theta, \phi)(t)$, and calculate it for the specified resolution of azimuth $\theta$ and elevation $\phi$ angles. The steering vector as shown in \Cref{eq:steering_vector} consideres the displacement of the robot in spherical coordinates $(d_i(t),\varphi_i(t),\xi_i(t))$, where $d_i(t)$ is the Euclidean displacement of node $i$ between $t_0$ and $t$.

\begin{equation}
    \mathbf{a}_{i}(\theta, \phi)(t) = 
    {e}^{-2\pi i \frac{d_{i}(t)}{\lambda}\sin\theta\sin\xi_i(t)\cos(\phi-\varphi_i(t))+\cos\xi_i(t)\cos\theta}
    \label{eq:steering_vector}
\end{equation}

As shown in \Cref{eq:bartlett}, the captured CSI data is then projected onto the various steering angles to compute the AoA profile of the received signal CSI along each virtual antenna element \cite{Stoica2005SpectralSignals}. The peaks in the resulting AoA profile correspond to the directions from which the strongest signals are arriving, thus enabling the estimation of relative bearing.

\begin{equation}
    F_{i j}(\theta,\phi)=\left|\sum_{t=t_{0}}^{t_{end}}h_{ij}^{2}(t)\mathbf{a}_{i}^2(\theta,\phi)(t)\right|^{2}
    \label{eq:bartlett}
\end{equation}

\subsection{Software pipeline architecture}
Our software pipeline is designed to provide a flexible, extendable, and realistic simulation framework that integrates MATLAB with Gazebo via ROS, using ROS1 Noetic and Gazebo Classic on Ubuntu 20.04. Central to the framework is the ROS Core, enabling communication between Gazebo and MATLAB through the ROS Toolbox. Control of the robots in Gazebo is identical to that of the real robots, as they interact with the same odometry publisher and trajectory subscriber. Moreover, MATLAB is used to spawn multiple robots in Gazebo and calculate the signal interaction between them.

MATLAB manages the publishers and subscribers and ensures the robots follow the right trajectory. 
The pipeline incorporates all CSI and odometry processing steps leading up to running the Bartlett's estimator and obtaining the estimated bearing angle. We store the simulated data in the same format as the real data. This enables our simulation framework to be able to process the real data, and the WSR toolbox to be able to process the simulated data for cross-referencing and validation. 

\section{Experiments}
\label{sec:experiments}

The real experiments implement the WSR toolbox on UP Squared Single-Board Computers (SBCs) with the Intel WL5300 NIC. We installed the software on Ubuntu 18.04, as the obsolete WiFi NIC used by the toolbox is not compatible with any newer Ubuntu versions. The robots are controlled using ROS1, specifically version Melodic Morenia, which is compatible with Ubuntu 18.04. The scripts to start the trajectory and the CSI collection on both robot nodes are activated over a Secure Socket Shell (SSH) connection from a third device. This connection is established using an additional USB WiFi transceiver, as the custom drivers for CSI collection impair the normal WiFi transmission capabilities of the NIC. 

This real experimental setup is paralleled in simulation using our framework described in \Cref{sec:sim_framework}. Similarly, ROS1 is used to match the setup of the real experiments. However, we opted for the latest version, ROS1 Noetic, on top of the latest compatible version of Ubuntu, which is version 20.04. This was installed on an HP EliteDesk 800 G2 MT (2015), based on an Intel\textsuperscript\textregistered Core\texttrademark i7-6700, quad-core 3.40GHz processor without dedicated graphics. The MATLAB interface performing the CSI signal calculations was deployed on a second machine, an HP ZBook Studio G5 (4QH37ES) (2018), based on an Intel\textsuperscript\textregistered Core\texttrademark i7-8750H CPU, hexa-core 2.20GHz with Quadro P1000 dedicated graphics. This second device runs MATLAB 2023b on top of Windows 11, with the ROS Toolbox Add-on installed to communicate with the Gazebo simulation. Despite distributing the computational load, we manually reduce the maximum update frequency of ROS to slow down the clock of Gazebo to be able to capture at 100Hz, matching that of the real set-up.
This simulated CSI is thus captured in simulation-time, and stored in the same format as the real experiment data. Given that the simulated and real robots contain identical publishers and subscribers on ROS, we can interact with real and simulated robots the same way.

All real experiments are performed at WiFi channel 108 with 30 subcarriers, with the same channel being used to calculate the simuated CSI. This channel contains a 20 MHz frequency range at 5.530-5.550 GHz, with a center frequency of 5.540 GHz. The center subcarriers are found to be the most accurate on the real hardware, of which subcarriers $12$ to $19$ are used to interpolate the true center subcarrier $15.5$.

\subsection{Stationary experiments}
We first conduct an experiment in which both nodes are kept stationary to examine the wireless channel behavior in the absence of movement.
This experiment was repeated three-times with a measurement interval of 100 Hz, for durations of 10, 30, 60, 120, and 300 s. In the simulation, the CSI was modeled using \(\textit{CFO}_{ij}(t) = \Delta f t\), where \(\Delta f\) is a constant representing the frequency offset between two unsynchronized antennas. This introduces a linear phase shift in the CSI with a slope \(\Delta f\), in addition, WGN is added to the simulation to replicate real-world noise conditions.

We calculate the value of \(h_{ij}^2\) using \Cref{eq:csi-multiplied}, which yields a consistent channel phase over time in both the simulation and the real experiments, despite the presence of minor noise. Given that the signal phase after CFO removal remains constant with minimal noise, we infer that CFO is the primary factor influencing the raw captured CSI, as described in \Cref{eq:cfo-channel-signal}. Assuming that CFO is a constant factor represented solely by \(\Delta f\), as implemented in the simulation, we can determine it from the real and simulated data. 
However, this approach did not yield the constant true phase, but rather periodic waves throughout the data, suggesting that the CSI in reality is more complex than just a constant offset. 
Unwrapping the phase data exposed a sinusoidal-like wave, further emphasizing the periodic nature observed. Based on these findings, we consider a simple but realistic model for the CFO as depicted in \Cref{eq:cfo-final}.

\begin{equation}
    \textit{CFO}_{ij}(t) = \Delta f t + c_1 \sin(c_2 t)
    \label{eq:cfo-final}
\end{equation}

\subsection{Mobile experiments}
\label{subsec:mob_exp}
We deploy two Turtlebot3 Waffle ground robots as shown in \Cref{fig:robot,fig:experiment-setup}, one as a mobile ($i$) and one as a static ($j$) node. 
The two nodes are $[3.858,0.929]$ m apart in $[x,y]$ directions in the reference frame of node $i$. This corresponds to a relative bearing of $2.8$ rad or $165$ degrees. The robots are deployed in an open environment on a flat 2D surface and with direct Line-of-Sight (LOS). The mobile node drives linear or circular trajectories. Moreover, data is gathered from both nodes while they are static in their initial position to gain additional insight into the CSI without movement used in \Cref{subsec:cfo}. We consider circular and linear trajectories of $30$ cm radius and $30\times2\pi$ cm length, respectively. Each trajectory is performed in 10 seconds at a velocity of about 0.188 m/s.
Bearing estimations require a trajectory of at least two times the size of the carrier signal wavelength, which equates to about 12cm at 5 GHz\cite{Jadhav2022ARobotics}.
The odometry and CSI data are both captured at sample rates of 100 Hz. This results in a datapoint every $0.0019$ m, well within the minimum sample threshold of $\frac{\lambda}{2}=0.0271$ m at 5.54 GHz  for successful bearing estimation \cite{Jadhav2022ToolboxRobotics}.
Three and ten repetitions are performed for each of our real and simulated experiments.
The CFO is added with $\Delta f=10.0$, $c_1=10000$, $c_2=200$, and with a uniformly distributed time difference between RX and TX of magnitude $\epsilon_t = 100$ ns. 

\begin{figure}[t]
\centering
\begin{subfigure}[b]{0.5\textwidth}
    \centering
    \includegraphics[width=1\linewidth]{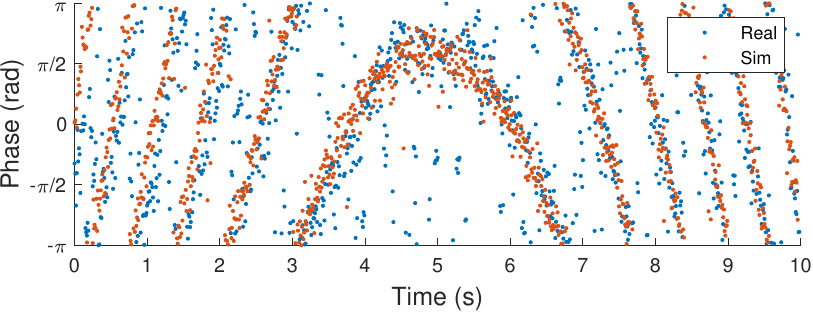}
    \caption{Real and simulation experiment, $30$ cm radius circular trajectory.}
    \label{fig:phase_circle}
\end{subfigure}
\\[\baselineskip]
\begin{subfigure}[b]{0.5\textwidth}
    \centering
    \includegraphics[width=1\linewidth]{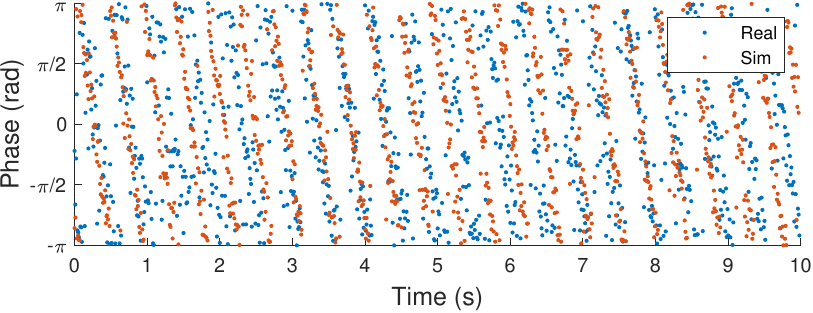}
    \caption{Real and simulation experiment, $30\times2\pi$ cm linear trajectory.}
    \label{fig:phase_line}
\end{subfigure}
\centering
\caption{Carrier signal phase obtained from both simulation and real experiments after cancellation of CFO.
Each data point corresponds to a sample taken at 100 Hz frequency. The real and simulation data match closely for both trajectories.}
\label{fig:phase_experiments}
\end{figure}

\begin{figure*}[t]
\centering
\begin{subfigure}[b]{0.475\textwidth}
    \centering
    \includegraphics[width=1\linewidth]{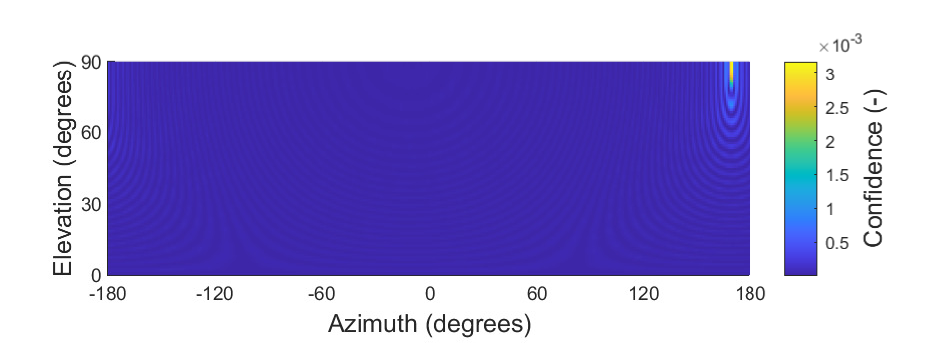}
    \caption{Simulation experiments $30$ cm circle radius trajectory.}
    \label{fig:aoa_circle_sim}
\end{subfigure}
\hfill
\begin{subfigure}[b]{0.475\textwidth}
    \centering
    \includegraphics[width=1\linewidth]{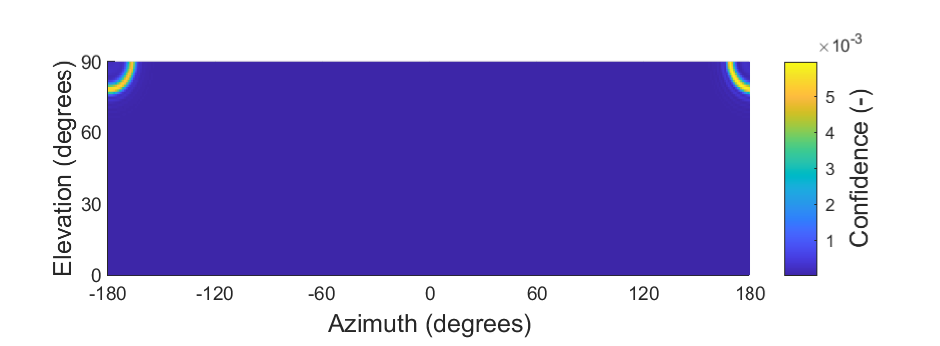}
    \caption{Simulation experiments $30\times2\pi$ cm linear trajectory.}
    \label{fig:aoa_line_sim}
\end{subfigure}
\\[\baselineskip]
\begin{subfigure}[b]{0.475\textwidth}
    \centering
    \includegraphics[width=1\linewidth]{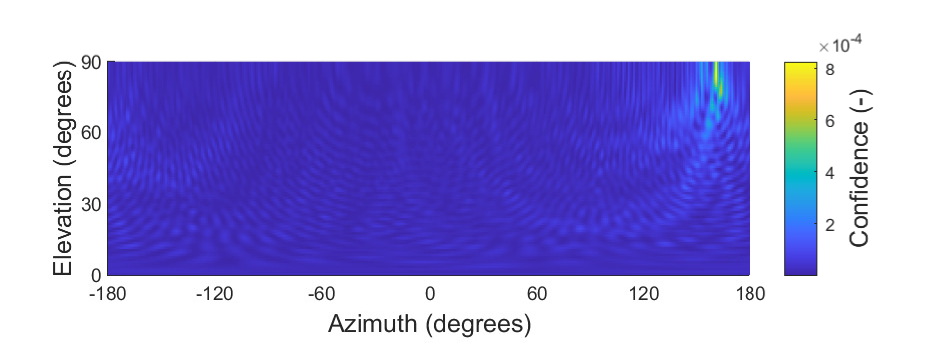}
    \caption{Real experiments $30$ cm circle radius trajectory.}
    \label{fig:aoa_circle_real}
\end{subfigure}
\hfill
\begin{subfigure}[b]{0.475\textwidth}
    \centering
    \includegraphics[width=1\linewidth]{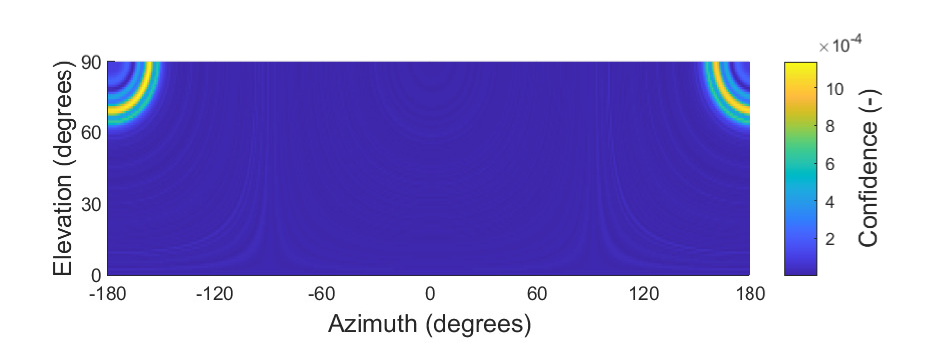}
    \caption{Real experiments $30\times2\pi$ cm linear trajectory.}
    \label{fig:aoa_line_real}
\end{subfigure}
\centering
\caption{The AoA profiles resulting from Bartlett's estimation for both circular (a,c) and linear (b,d) trajectories. We perform 10 iterations on simulation (a,b) and 3 iterations on real (c,d) experiments. The AoA profiles shown here are the average of experiment repetitions in each case. While the circular trajectory results in a unique bearing estimate, the linear trajectory returns two mirrored possibilities. Real AoA profiles demonstrate more noise compared to those obtained from simulation.}
\label{fig:aoa_results}
\end{figure*}

\section{Results \& discussion}
\label{sec:results}

\Cref{fig:phase_experiments} shows the CSI phase data after canceling the CFO in the real and simulated signals as described in \Cref{eq:cfo-final}. The uniformly distributed random term of $\epsilon_t$ as described in \Cref{subsec:mob_exp} account for the noise in the simulated data. 

\Cref{fig:aoa_results} demostrates that the real and simulated experiments can accurately estimate the bearing angle when a circular trajectory are used. The real experiments show slightly higher irregularity in the grating lobes caused by factors not equivalently replicated by the simulation. Additional peaks in the AoA profiles of real experiments can result from multipath signals. Moreover, the extra noise visible in \Cref{fig:aoa_circle_real,fig:aoa_line_real} can also be caused by errors in the odometry and location, where the simulation uses accurate odometry.
These factors cause a slightly lower confidence in the AoA peak observed in real experiments. 

We observe mirrored bearing estimations from linear trajectories in both the simulation and the real experiments. This phenomenon occurs when the antenna array contains ambiguity in the form of two equally likely bearing estimations mirrored over the linear trajectory path \cite{Jadhav2022ARobotics}. 
In order to obtain accurate azimuth angle estimations, the trajectory should ideally be circular on a plane or a line that contains the transmitter, confirming findings reported in \cite{Kumar2014AccurateCost, Jadhav2022ARobotics}.

The simulated data is based on the theoretical carrier signal equation using accurate odometry, which is also used in the steering vector for Bartlett's estimation. This accuracy in the odometry information potentially gives the simulation an edge in accuracy over the real experiments. 
In future work, we plan to leverage our simulation framework to enhance the modeling of real-world multi-paths propagation of signals, and a more realistic method to add noise to the odometry. This can include a more detailed model-based approach like the one considered in this work, or a data-driven approach where a variety of environmental settings and scenarios are considered in the real world in order to train a neural network that can replicate the corresponding noise and multipath effects on the theoretical data.


\section{Conclusion}
\label{sec:conclusion}

We present a simulation framework that leverages Gazebo and Matlab to replicate the operation of the WSR toolbox delivering WiFi-CSI processing as a sensing technology in a realistic environment. We validate our simulation framework against real experiments in a constant laboratory environment and by using a set of basic trajectories that create a virtual antenna array. Signal properties, including CFO, STO, and noise, present in real experiments are replicated using a systematic approach in simulation. Assessment of our experimental and simulation results demonstrate a slightly more confident bearing estimation in simulation compared to real experiments due to less noisy conditions. The mobile receiver node is found to accurately estimate the bearing of the static transmitter in both the real and simulation experiments.
We conclude that our single-path-based CSI simulation closely matches the multi-path real experiment CSI in LOS scenarios. This work effectively extends the suite of available sensors in simulation environments for localization purposes, going beyond camera and LiDAR systems.
In our future work, we plan to update the hardware systems for capturing CSI data to more commonly available devices. By integrating the CSI collection software into a ROS package where the data becomes available over ROS, we will aim to ensure that the software framework is easily deployable to facilitate seamless experimentation and development between real and simulation experiments. 

\newpage

\bibliographystyle{IEEEtran}
\bibliography{./references}

\end{document}